\documentclass[dea, biber]{now-journal}

\usepackage{epstopdf}
\usepackage{enumitem}

\usepackage{multirow}

\newcommand{\mcal}{\mathcal}
\newcommand{\mb}{\mathbf}

\title{Federated Analytics: A survey}

\author{Ahmed Roushdy Elkordy}
\affil{University of Southern California, USA}
\author[1]{Yahya H. Ezzeldin}
\author{Shanshan Han}
\affil{University of California, Irvine, USA}
\author{Shantanu Sharma}
\affil{New Jersey Institute of Technology, USA}
\author{Chaoyang He}
\affil{FedML Inc.}
\author[2]{Sharad Mehrotra}
\author[1]{Salman Avestimehr}

\newcommand{\blue}[1]{\textcolor{black}{#1}}
\newcommand{\blueP}{\color{black}}

\articledatabox{ISSN 2161-1823; DOI 10.1561/103.00000003\\
\copyright \ ~}


\begin{document}

\begin{abstract}
Federated analytics (FA) is a privacy-preserving framework for computing data analytics over multiple remote parties (e.g., mobile devices) or silo-ed institutional entities (e.g., hospitals, banks) without sharing the data among parties. Motivated by the practical use cases of federated analytics, we follow a systematic discussion on federated analytics in this article. In particular, we discuss the unique characteristics of federated analytics and how it differs from federated learning. We also explore a wide range of FA queries and discuss various existing solutions and potential use case applications for different FA queries. 
\end{abstract}
\section{Introduction}
Federated Analytics (FA) is a paradigm for collaboratively extracting insights from distributed data that is owned by multiple parties (\textit{e}.\textit{g}., individual mobile devices or institutional organizations) under the coordination of a central entity (\textit{e}.\textit{g}., a service provider) without any of the raw data leaving their local parties or revealing information beyond the targeted insights. The core principles of this paradigm allow breaking the limitations for deriving analytics from limited centralized data, in terms of privacy concerns and operational costs. In the last decade, federated learning~\citep{kairouz2021advances}, a closely related area to federated analytics, has received significant interest both in academic and industry domains. Recently, the research community is extending federation beyond learning settings to address more generalized analytics questions. In this work, we summarize the diversity of questions within federated analytics and highlight research problems that can have significant theoretical and practical interests.

The term federated analytics was first coined by Google in 2020\footnote{https://ai.googleblog.com/2020/05/federated-analytics-collaborative-data.html} to represent ``collaborative data science without data collection''. It was first explored in support of federated learning as a way for Google engineers to evaluate the quality of the learned machine learning models against real-world data. Beyond model evaluation, FA implementations have expanded to other applications with the flagship application being the discovery of popular elements across devices, \textit{e}.\textit{g}., popular out-of-dictionary words~\citep{zhu2020federated} or most popular songs recognized by phones. 
In these FA applications, the key challenge was to develop protocols that are efficient at scale while taking into account the limited communication bandwidth, as well as preserving the privacy of the participating parties.

Even with the success of these initial FA solutions and the recent interest in this collaborative paradigm, there is, unfortunately, no clear definition for what constitutes federated analytics, what kind of interesting analytical questions it can answer, and what are the possible real-world domains that can benefit from its applications. Very recent summarizing efforts in federated analytics have focused on queries of interest to particular domain applications such as video analytics~\citep{video_analytics_FA_survey}. However, there exists a wide range of other queries that can be supported (and are of interest) in an FA system. Summarizing these different query classes and the potential approaches for answering them in federated analytics provides a great starting point for new researchers in this areas as well as the future development of generalized solutions for serving these queries within an FA system.

\begin{figure} 
\centering
\includegraphics[width=0.97\textwidth]{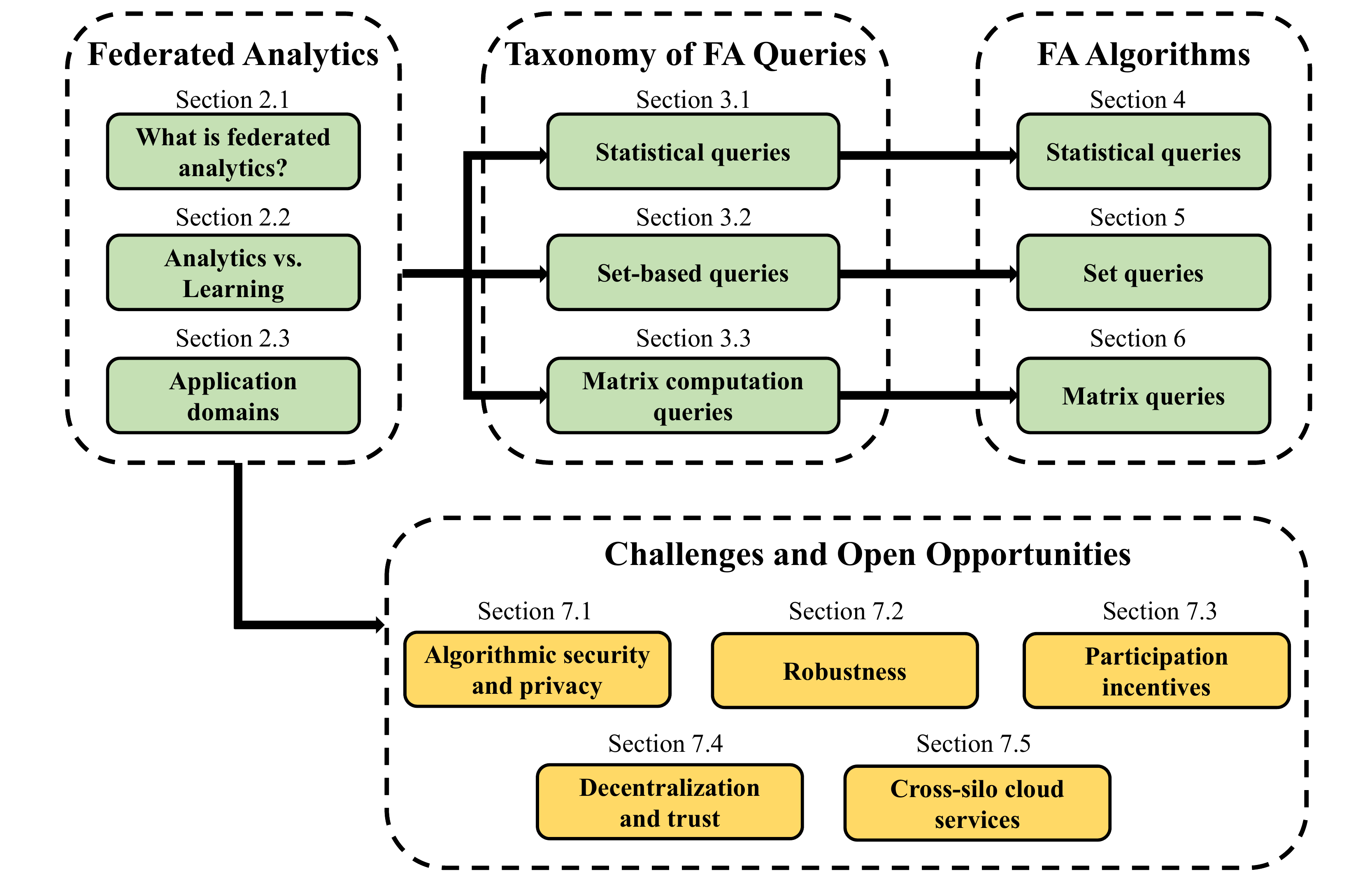}
\caption{The schematic structure of federated analytics and the relationship between different sections. The body of this survey mainly contains the fundamentals of federated analytics, a taxonomy of different queries of federated analytics, federated analytics algorithms, applications, and discussions of challenges and opportunities in federated analytics in the presence of cloud-based services.}
\label{fig:structure}
\end{figure}

\blue{This paper aims to provide an introductory guide to federated analytics as follows (Figure~\ref{fig:structure}). We first define federated analytics and how it relates to the more well-studied field of federated learning. Next, we provide a taxonomy of typical data analysis queries of interest in federated analytics and where they can find use in different domains. For the presented queries, we also discuss different existing approaches in the literature for addressing them. Finally, we discuss different challenges and opportunities within the federated analytics framework and discuss potential solutions for addressing these challenges and open directions. These open questions provide starting points for expanding and developing more practical scenarios in federated analytics, where research efforts are still needed.}

\section{What is federated analytics?}\label{sec:FA_definition_sec}
{In federated analytics, there is typically a central querier (the question asker) who wants to learn some property or answer a question based on data distributed across different clients (\textit{i}.\textit{e}., parties). Each of these clients owns a subset of the data, representing their local dataset. We will refer to these parties as clients or data owners interchangeably throughout this survey.


From a generalized perspective, \textbf{federated analytics} can be defined as a setting for data analysis where a querier wishes to answer a data analysis query through the collaboration of multiple data owners (clients) that own their local raw data. The raw data is not exchanged or transmitted, but instead, intermediate query replies that are meant for aggregation at the querier are transferred to answer the intended query.

In particular, from this generalized view, the goal of federated analytics is for a central querier to answer the following query $Q$
\begin{equation}\label{eq:query_def}
    Q(\mcal{D}) = F_\omega(\mcal{D}_1, \mcal{D}_2, \cdots, \mcal{D}_N).
\end{equation}

\begin{figure} 
\centering
\includegraphics[width=0.57\textwidth]{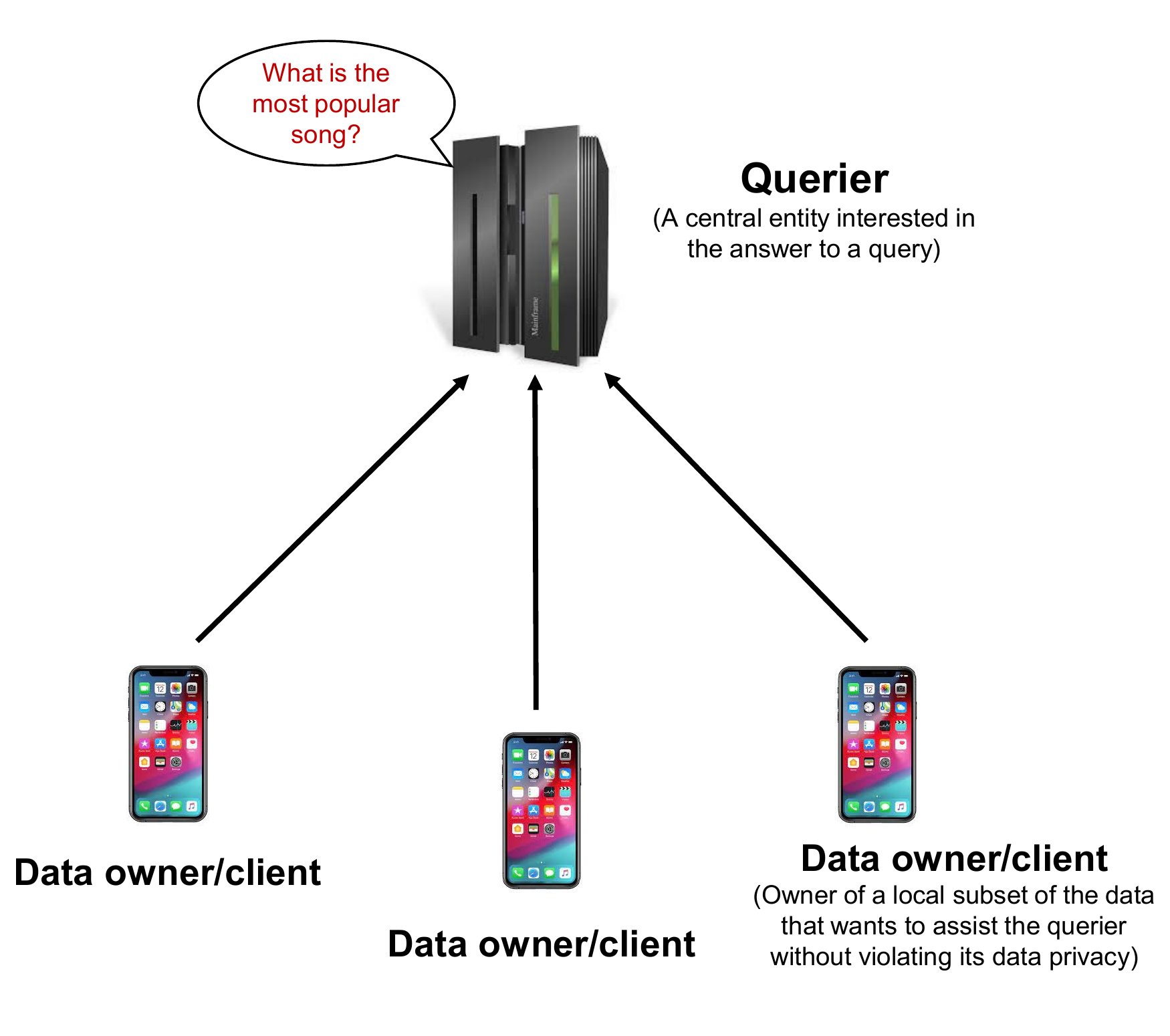}
\caption{An example federated analytics setting where a \textbf{querier} is discovering the most popular song in the collective datasets at the clients, where each client is a \textbf{data owner} of its local subset. To preserve the privacy of the clients' data the system seeks to answer the query distributively with only focused replies being sent back to the querier.}
\label{fig:system_model}
\end{figure}

Here $\mcal{D} = \{\mcal{D}_i\}_{i=1}^m$ is the private datasets at the $N$ data owners, and $F_\omega$ is the (potentially parameterized) function on the data describing the target query. For instance, given a pre-trained machine learning classification model parameterized by $\omega$, the basic federated analytics query to test {\it the accuracy of the model $\omega$} on the distributed datasets can be represented by the following query:
\begin{align}
Q_\omega(\mcal{D}) &= Acc(\omega; \{\mcal{D}_1, \mcal{D}_2, \cdots, \mcal{D}_N\}) \nonumber\\
  &= \sum_{i=1}^N \frac{|\mcal{D}_i|}{\sum_{i=1}^N |\mcal{D}_i|} Acc(\omega; \mcal{D}_i),
\end{align}
with the query answer being the weighted average of each party's local test accuracy $Acc(\omega;\mcal{D}_i)$. To compute the local accuracy, each party applies the model to its local labeled dataset and computes the local ratio of correct classifications.}

\subsection{Federated learning vs. federated analytics}
Federated analytics is very similar to federated learning~\citep{kairouz2021advances} in the fact that both require collaborative use of distributed data without collecting the raw data at a centralized location. 
However, while federated learning, as a branch of distributed optimization, is about training machine learning models at the edge and aggregating learning outcomes back into the federated learning model, {\blueP federated analytics is more generalized to include applying basic data science methods for data analysis but also includes optimization-based questions such as federated learning. Thus from a generalized perspective using the formulation of~\eqref{eq:query_def}, federated learning can be viewed as a complex federated analytics query on the distributed datasets when the function $F_\omega$ is the following optimization empirical risk minimization problem:
\begin{equation}\label{eq:fl_as_fa}
F_\omega(\mcal{D}_1, \mcal{D}_2, \cdots, \mcal{D}_N) = {\rm arg}\min_\mb{w} \sum_{i=1}^N \sum_{\mb{x} \in \mcal{D}_i}\ell(\mb{w};\mb{x}).    
\end{equation}
The analytics branch of federated learning has been extensively studied in recent years~\citep{kairouz2021advances}, while algorithms and approaches for basic data science queries have not seen similar exploration, even though they are critical to service federated learning models. In fact, one of the first application examples of non-learning queries in federated analytics is strongly coupled with federated learning, where engineers at Google wanted to evaluate the inference performance (\textit{e}.\textit{g}. in terms of accuracy) of trained federated learning models against real-world data not available at the data centers.

Thus, in the remainder of the paper, we limit our attention to simple federated analytics queries that would not require optimization when solved in a centralized scenario, in contrast to the federated learning branch which would require optimization of parameters to solve in a centralized setting.
Following this distinction, examples of simple queries for federated analytics include questions of the form: what is the mean or median value of a function applied on the distributed data; while federated learning would be confined to learning a parameterized function such as: what is the best model that maps features $\mb{x}$ to target variable $\mb{y}$. In fact, each round of federated learning invokes the simplest question in federated analytics after local training: {\it what is the sum of vectors (gradient updates) stored at the participating clients?}.}



\begin{figure} 
\centering
\includegraphics[width=0.95\textwidth]{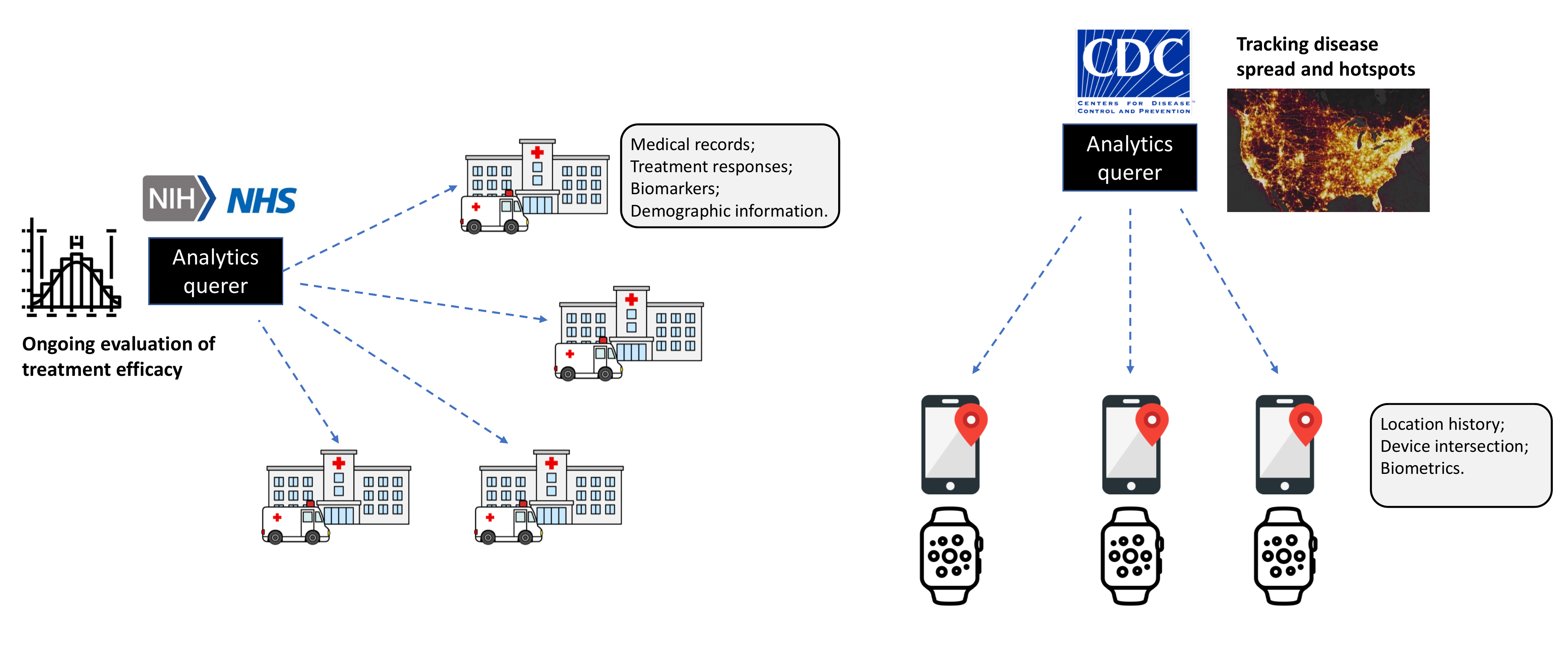}
\caption{Examples of federated analytics applications in the healthcare domain.
}
\label{fig:examples}
\end{figure}
\subsection{Applications for federated analytics}
 We, next, discuss several canonical domains that benefit significantly from applying federated analytics. Figure~\ref{fig:examples} highlights a number of these applications of federated analytics in the healthcare domain.

\begin{itemize}
    \item \textbf{Evaluation analytics for machine learning models.} The poster application that started garnering interest in federated analytics was the collaborative evaluation of the quality of trained machine learning models. For instance, Google uses federated analytics to evaluate the accuracy of Gboard next-word prediction models by using captured data from users' typing activities on their phones. Similar to accuracy evaluation, federated analytics can also be used to compute other evaluation metrics of the trained machine learning models, \textit{e}.\textit{g}., model robustness to unseen distributions/users as well as the fairness to different demographic groups~\citep{ezzeldin2021fairfed} (for example, how different is the performance of an image tagging application to photos from the black vs white communities).
    
    \item \textbf{Analytics for medical studies and precision healthcare.} A key ingredient for realizing the full promise of precision medicine is allowing research analytics and diagnostics on large amounts of medical data that are not typically available through traditional medical research procedures. This kind of information can originate from data collected at medical institutions (\textit{e}.\textit{g}., the efficacy of applied treatments and onset symptoms associated with a diagnosis) to individual personal data such as location history of individuals for contact tracing (\textit{e}.\textit{g}. during COVID-19), or mental health studies based on bio-markers. Enabling these gains from big medical data is challenged by the legal and regulatory barriers for privacy that make collecting patient-level data outside a healthcare provider complex and time-consuming.
    
    \item \textbf{Guiding advertisement tactics.} Advertisers are keen to know whether their ads are attractive to their potential customers. For example, in the case of video ads, they would like to collect summary ads viewership data from users to understand the effectiveness of their advertisement concepts as well as guide future advertisement expenditure.
\end{itemize}

The aforementioned domains can make use of a large number of simple federated analytic metrics beyond the promise of federated learning models. In the following section, we give a taxonomy of different federated analytics queries and highlight to the reader some of their potential use cases in the discussed application domains.

\section{A taxonomy of federated analytics queries}\label{sec:query_types}
\begin{figure} 
\centering
\includegraphics[width=0.97\textwidth]{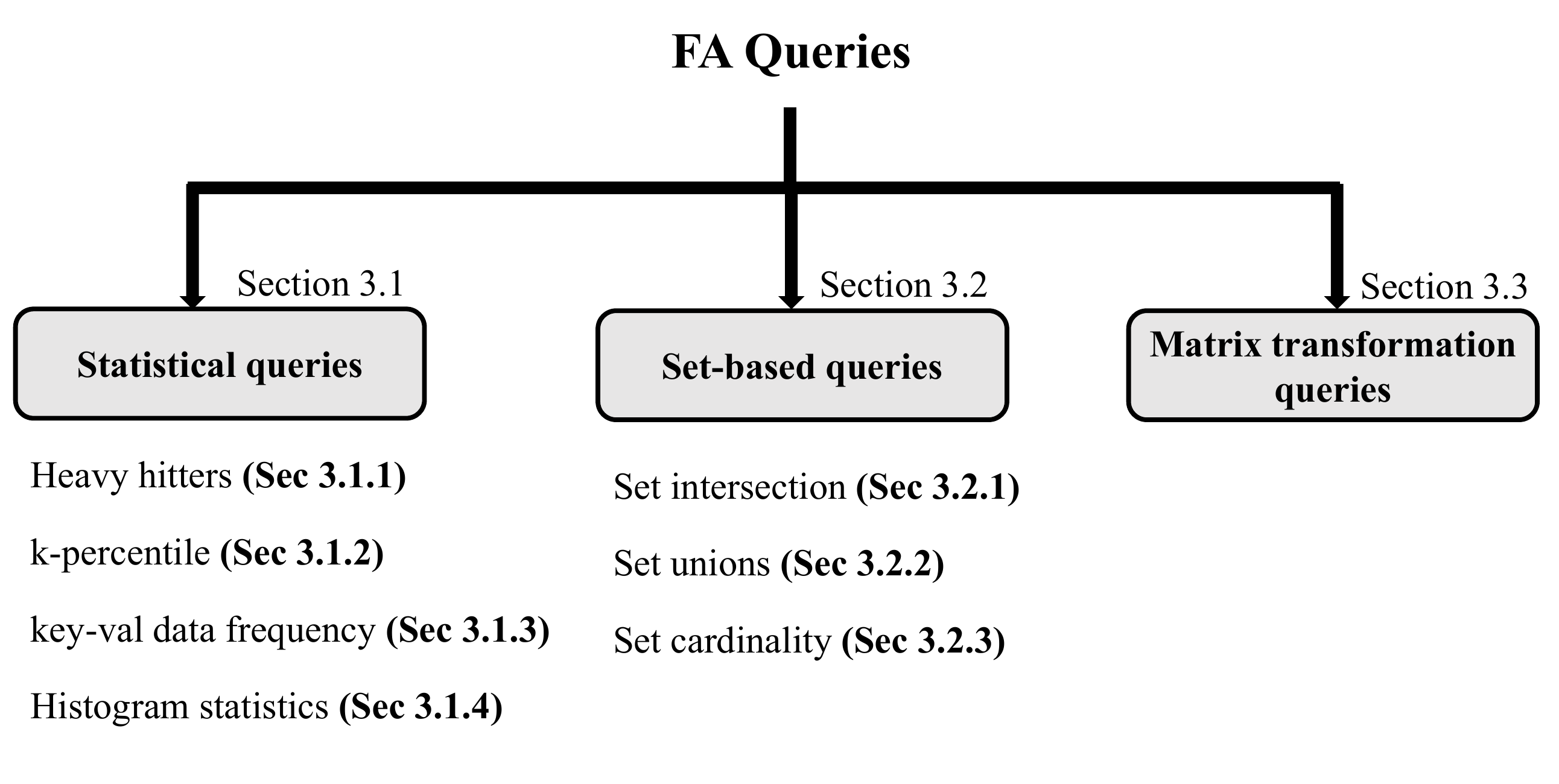}
\vspace{-1em}
\caption{A taxonomy of federated analytics queries presented in Section~\ref{sec:query_types}.}
\label{fig:query_taxonomy_figure}
\end{figure}

\blue{As described in Section~\ref{sec:FA_definition_sec}, a federated analytics query is a general class that encompasses any question by a querer on distributed private datasets. However, from this general class of queries, there exist a number of queries that find greater exposure in different application domains and are explored more deeply in the literature. We can divide these queries of interest into three main categories: 1)  Statistical testing queries, 2)  Set queries, and 3) Matrix transformation queries. 
The statistical testing category includes different data science queries that aim to discover key statistical properties of the distributed private data. Examples of such queries would be the estimation of the mean median, heavy hitters, key-valued data frequencies, hypothesis testing, \ldots, etc. The set queries, on the other hand, include analytics for discovering data associations such as set intersection, set union, and intersection cardinality. Matrix transformation queries include but are not limited to operations such as dimensionality reduction using methods such as principal component analysis, and projections. In this section, we formally define the most popular queries in each of the aforementioned query types and present some of their real-world applications. Figure~\ref{fig:query_taxonomy_figure} summarizes the queries presented in the remainder of this section.}

\subsection{Statistical testing} 
We focus on four key statistical queries that have a wide variety of real-world applications in different domains, such as health, business, and user experience. For each of these statistical queries, we give its mathematical definition, followed by one of its main applications. We discuss some existing solutions in Section \ref{sec-related-works-stastics}.   We start by first assuming  having a set $ \mathcal{D} = \{\mathcal{D}_1, \dots, \mathcal{D}_N\}$ of $N$ datasets, where each  dataset $\mathcal{D}_i = \{x^i_1, \dots, x^i_{n_i}\} $ consists of $n_i$ data points and is owned solely  by one  distributed node, \textit{i}.\textit{e}., an FA client.  
\subsubsection{Heavy hitters}
The objective of the heavy hitter problem is to construct a succinct histogram of the elements across the $N$ parties datasets that contains only the most popular (heavy-hitter) elements; other elements are treated as if appearing with zero frequency. Typically, an element is denoted a heavy-hitter if its frequency in the distributed dataset is greater than or equal to a fraction $\phi$ of the dataset size. Formally the goal of the query is to return the following:

\begin{align}
    &Q(\mcal{D}) = \{(x, {\rm \mathit{freq}}(x)) | x \in \mcal{D}_{\rm HH}\} \nonumber \\
    &\text{where:}\quad  
    \mcal{D}_{\rm HH} = \left\{x \left| x \in \bigcup_{i=1}^N \mathcal{D}_i,\ \  {\rm \mathit{freq}}(x) \geq \phi |\mcal{D}|\right.\right\}.
\end{align}

Note that the heavy-hitters problem is closely related to another succinct histogram problem formulation, the top-$K$ problem, where the goal is to find a succinct histogram with the $K$ most frequent elements instead of all elements exceeding a threshold. If we target the top-1, this translates to the well-known \textit{mode} statistic of the dataset. 

\medskip

\noindent {\bf Application (User Experience). } One popular application of heavy hitters is to learn trendy out-of-dictionary words generated by users' devices. Learning trendy words is of high interest to service providers as it allows them to improve the service they provide to their users. These services could be  the autocomplete feature  in  smart keyboards,   or a powerful    advertisement  engine that could   leverage  the   current public taste of people for more effective advertisement. A similar application is to learn the out-of-dictionary words, which can be used to improve the smart keyboard spell-auto-correction feature by adding such words to the keyboard’s dictionary. Apple has already used differential privacy to protect the privacy of users' input data while collecting the top frequent emojis by users \citep{apple}. Similarly, Google has also proposed another differential privacy (DP) method to collect the out-of-dictionary words \citep{pp}.

\subsubsection{k-percentile element}{$k$-percentile element} 
In the $k$-th percentile statistical query problem, the objective is to find the smallest element that is greater than $k$ percent of the overall dataset available at the participating distributed nodes. This statistical query problem can be formalized as follows.        
Assuming the entries of the datasets in $\mathcal{D}$ are non-categorical values (\textit{i}.\textit{e}., numerical values), then  by denoting $\mathcal{D}^s$ to be the  non-decreasing  sorted set of the  elements   of  $\bigcup_{i =1}^N \mathcal{D}_i$,   the $k$-percentile element $x_k$  in this distributed parties datasets $\mcal{D} = \{\mcal{D}_1, \mcal{D}_2, \cdots, \mcal{D}_N\}$ is  given by
\begin{equation}
    Q(\mcal{D}) = x_k = x\text{ such that } \mathit{rank}_{\mathcal{D}^s} (x) = k\times|\mcal{D}^s|, 
\end{equation}
where $\mathit{rank}_\mathcal{D}^s (x)$ is the order of element $x$ in the dataset $\mathcal{D}^s$. An example of $k$-percentile values is the \textit{median}, where $k$ is $0.5$.

\medskip

\noindent\textbf{Application (Business).}
 It is well-known that the median is a more robust metric to represent central tendency compared to the mean, which is more sensitive to outliers. Hence, it is more useful in business use cases to assess different components such as company salaries. For instance, a possible application for federated median computation is for an authority to compute the median salary (or any other percentile) of all employees in a set of companies without revealing the exact salaries of the employees or which companies they belong to.

 
\subsubsection{Key-valued data}
The  Key-valued data  is a statistical query problem in which each data point is represented by a key (\textit{e}.\textit{g}., identifier) and value associated with this key, while  the objective is to learn the frequency of each key and the mean (or aggregate) of the values that appear paired with this particular key. To formalize the objective, we assume  that the dataset $\mathcal{D}_i$, for $i \in [N]$ is a key-valued dataset such that 
$\mathcal{D}_i = \left\{  x_j^i \big |   x_j^i = (k_j^i,v_j^i), \; \forall j \in [n_i]\right\} $. The objective is to find the following 

\begin{equation}
   Q(\mcal{D}) =  \left\{ \left(\mathit{freq}(k_i), \; \frac{1}{|\mathit{freq}(k_i)|} \sum_{v_j: (k_i, v_j) \in \mathcal{D}}  v_j \right), \forall k_i \in  \mathcal{D} \right \}.
\end{equation}

\noindent\textbf{Application (Business).} 
A possible application can be in the business market, where the objective is to privately learn the distribution of the stocks and the investment amount of each stock from the private data of the investors. Specifically,   in this stock market application,   the key represents the stocks while the value represents the amount that a person invests in a given stock. The statistical query goal takes place when an analyst wants to learn how many agents invest in each stock (\textit{e}.\textit{g}., frequency distribution stocks) and the amount invested in each stock (\textit{e}.\textit{g}., average or aggregate amount)  without collecting any private data which can cause a breach to their privacy.

\subsubsection{Histogram-based statistics}
This can be considered a special case of the key-valued data problem, where the objective is to learn only the frequency of each key. 

\noindent\textbf{Application (User experience). }  One real-world application of histogram-based statistics is the  Now Playing feature on Google’s Pixel phones \citep{NowPlaying}.   This feature uses an on-device database of song fingerprints to show users what song is playing in the surrounding room without an internet connection. The one-device database includes the most frequently recognized songs, which are maintained and updated by Google to ensure that the database contains only popular songs. The way it works is that on each phone, the Now Playing application computes the recognition rate (value) for each song (key) in its Now Playing History. Once the phone is plugged in and connected to WiFi, the users encrypt the rate of the songs and send them to the Google servers so that they can only compute a histogram distribution of all song counts. This allows Google to replace the less popular songs in the database with the more popular ones.

\subsection{Private set queries} 
The distributed private set queries class can be broadly clustered into three different categories; distributed sets intersection, distributed sets union, and distributed cardinality computation. The main goal of this analytic problem is to compute these queries in a way that protects the privacy of the data owners being queried.    Similar to the statistical testing class, we consider having $N$ parties where each party $i$ has a dataset $\mathcal{D}_i$ of $n_i$ unique and private data points. Some of the existing solutions to set queries are presented in Section \ref{sec_related-works-sets}.    
\subsubsection{ Private Set Intersection}
The private set intersection (PSI) is a private set query problem that has a wide range of applications with the objective of computing the intersection between the sets owned by the different clients and nothing beyond that. This   query is formally  given as follows 
\begin{equation}
   Q(\mcal{D}) = \bigcap_{i=1}^N\mathcal{D}_i.
\end{equation}
\noindent\textbf{Application (Business).}
One famous application of PSI in the two-party setting  is the online-to-offline advertisement conversion \citep{51026} in which a company would like to know how much of its revenue can be  attributed to an online advertisement in order to assess the future payment it spends on a paid ad (\textit{e}.\textit{g}., Facebook ad). On the other hand,  the advertising company wants to know how successful its advertising campaign is. In this setting,  the advertising companies have a database of the users and their status, whether they saw the ad or not, while the company knows the users who purchased their products as well as the amount they spent on their purchases. In other words,  the data needed to compute these statistics are split across the two parties. In this setting,  the two parties are typically unwilling to share their customers' data to protect the privacy of their business and their customers, but both parties would want to collaboratively learn how many users both saw an ad and made a corresponding purchase, as well as the amount of money those users spent on the company's products.  

\subsubsection{Union}
Similar to private set intersection, the goal is to privately  evaluate the union  of the input sets of two or more  parties  privately without revealing anything about the
sets beyond the union. 
This objective can be formally given by 
\begin{equation}
   Q(\mcal{D}) = \bigcup_{i=1}^N\mathcal{D}_i.
\end{equation}

\noindent\textbf{Application (Security). } 
One popular application is  
risk assessment and management \citep{ramanathan2020blag}. The goal of this application is to aggregate the blacklists from different parties and across various attack types. This could help in improving the individual
blacklists in identifying malicious sources.
\subsubsection{Cardinality}
The goal of this problem is to learn the cardinality of the intersection of the data set of multiple parties in a private manner, which can formally be given as follows \begin{equation}
   Q(\mcal{D}) = |\bigcap_{i=1}^N\mathcal{D}_i|.
\end{equation}

\noindent\textbf{Application (Public Safety)}    One popular real-world application of PSI cardinality is the   CSAM Detection system used by apple ``Apple for Child Sexual Abuse Material (CSAM)''. The main goal is to identify and report iCloud users who store known Child Sexual Abuse Material (CSAM) in their iCloud Photos accounts. The way it works is that intersection cardinality testing is carried on between a known database of CSAM images and individual iCloud users. When the cordiality of intersection exceeds a predefined threshold,      Apple can provide relevant information to the National Center for Missing and Exploited Children (NCMEC).

\subsection{Matrix transformations}

Singular value decomposition (SVD) is one of the most popular matrix operations that have a wide range of applications in either data analytics or machine learning. The main objective of this problem is to compute SVD over a set of distributed data without collecting any raw data or breaching the privacy of the data owners. This problem can be formally defined  as follows: assume there are $n$ parties, and each party $i$  has a private data matrix $\mathbf{D}_i \in \mathrm{R}^{m \times n_i}$. The $n$  parties would like to compute the SVD jointly on the combined dataset $\mathbf{D} = [\mathbf{D}_1, \dots, \mathbf{D}_n]$, where $\mathbf{D}\in \mathrm{R}^{m \times n}$ and $n = \sum_{i=1}^n n_i$. The  private computation of SVD on the  combined dataset takes the following form
\begin{equation} \label{eq-svd}
    Q(\mcal{D}) = \mathbf{U} \Sigma [\mathbf{v}_1^T, \dots, \mathbf{v}_n^T] 
\end{equation}
where $\mathbf{U}$ and $\Sigma$ are shared across all the parties, while  $\mathbf{V}_i$, $\forall i \in [n]$, is  kept secret by party $i$ and never shared with any other parties. From \eqref{eq-svd}, each node $i$ can get its SVD by using the shared matrices  $\mathbf{U}$ and $\Sigma$, and the secret matrix $\mathbf{V}_i$ as $\mathbf{D}_i = \mathbf{U} \Sigma \mathbf{V}_i$.

Another variant of SVD called Funk-SVD is applied to the sparse rating matrix used in the recommendation systems  \citep{chai2020secure} such that it composes the sparse matrix into two embedding matrices that can be used to predict the missing rating in the rating matrix.  

\noindent\textbf{Application (Machine Learning).}
SVD is an essential building block in many studies and applications, such as principal component analysis (PCA). PCA  is used to reduce the feature space of the data used in machine learning. Reducing dimensionality in statistical machine learning can prevent the model from overfitting, which reduces the ability of the model to generalize beyond the examples in the training set. One challenge of performing PCA in a distributed setting is having the data distributed across multiple nodes while collecting and gathering the data is prevented by the law  (\textit{e}.\textit{g}.,  GDPR \citep{voigt2017eu}). 
We discuss some existing solutions for the matrix transformation query  in Section \ref{sec-related-works-stastics}. 



\section{Existing solutions to statistical testing queries}\label{sec-related-works-stastics}
\begin{table}
\resizebox{0.98\textwidth}{!}{%
\begin{tabular}{|l|l|l|l|}
\hline
 \textbf{Query}             & \begin{tabular}[c]{@{}l@{}} \textbf{Privacy} \\ \textbf{technique} \end{tabular} & \textbf{Related works}                                                                                                                                                                             & \begin{tabular}[c]{@{}l@{}} \textbf{Noisy} \\ \textbf{response} \end{tabular} \\ \hline
\multirow{4}{*}{Heavy hitters} & Non-private                         & \citep{charikar2004finding,cormode2003finding}                                                                                                                                           & No             \\ \cline{2-4} 
                               & DP                           & \begin{tabular}[c]{@{}l@{}}\citep{hsu2012distributed,bassily2015local,bassily2017practical} \\ \citep{apple,acharya2019hadamard,acharya2019communication,zhu2020federated}\end{tabular} & Yes            \\ \cline{2-4} 
                               & MPC                          & \citep{9519492}                                                                                                                                                                          & No             \\ \cline{2-4} 
                               & DP + MPC                     & \citep{DP-MPC}                                                                                                                                                                           & Yes            \\ \hline
\multirow{3}{*}{Median}        & Non-private                          &  \citep{iutzeler2017distributed}                                                                                                                                                                                          &         No       \\ \cline{2-4} 
                               & DP                           &   \citep{boehler2022secure,bohler2020secure+}                                                                                                                                                                                      &            Yes    \\ \cline{2-4} 
                               & MPC                          &       \citep{aggarwal2010secure,goldreich2019play,tueno2020secure}                                                                                                                                                                                      &   No             \\ \hline
Key-valued data                & DP                           &   \citep{8835348,gu2020pckv}                                                                                                                                                                                         &     Yes           \\ \hline
\end{tabular}}
\vspace{0.5em}
\caption{Taxonomy of the privacy-preserving techniques used in the statistical query.}
\end{table}

A taxonomy of the privacy-preserving techniques used for the statistical  testing queries is given in Table 1. We consider different variants of privacy-preserving techniques represented by differential privacy (DP),  secure multi-party computing  (MPC), and  a combination of  DP with MPC.

\subsection{Heavy hitters}
The heavy hitter problem has been well studied in the literature either in the centralized setting with no privacy requirements where  the data is already collected and stored at a central server or in a distributed federated setting where the queerer wishes to learn the “heavy hitters” in the  clients’ data while guaranteeing the privacy of each contributing client at minimal computation/communication costs \citep{charikar2004finding,cormode2003finding, charikar2004finding,cormode2003finding,hsu2012distributed,bassily2015local,bassily2017practical,apple,fanti2015building,acharya2019hadamard,acharya2019communication,zhu2020federated}). 

\subsubsection{Non-private centralized setting}   In the non-private centralized setting, the main objective is to develop efficient heavy hitters algorithms with low storage requirements and provable error bound. The low storage requirement is of significant importance when dealing with a large online data stream that memory-intensive solutions such as sorting
the stream or keeping a counter for each distinct element are infeasible  (\textit{e}.\textit{g}., \citep{charikar2004finding,cormode2003finding}).
The work in \citep{charikar2004finding}  proposes an approximate heavy hitter algorithm that is memory efficient with proven theoretical error bound. The algorithm is based on  sketch counting that relies on using a set of hashes that map each element in the data stream to  different bins, such that when running the sketch counting algorithm along with a max-heap data structure, the algorithm  can find the $k$ heavy hitters in a stream of $d$ unique items with storage cost logarithmic in $d$ (\textit{e}.\textit{g}., $O(K \log d)$)  instead of being linear in $d$.


\subsubsection{Private distributed setting}   
There is a rich body of works on private heavy hitters and frequency estimation in the distributed setting while ensuring users' privacy by leveraging  DP \citep{hsu2012distributed,bassily2015local,bassily2017practical,apple,acharya2019hadamard,acharya2019communication,zhu2020federated}, MPC  \citep{9519492}, or combine DP with MPC \citep{DP-MPC}.

\noindent \textbf{Heavy hitters with differential privacy.} Researchers have proposed multiple \textit{efficient} private heavy hitter algorithms that have a computation time, communication cost, and storage cost polynomial in $n$ (number of users) and logarithmic in $d$,  $log(d)$, where $d$ is the size of the data universe (dictionary of the data points to check). \citep{hsu2012distributed} proposed several efficient $(\epsilon, \delta)$-differentially private algorithms for the heavy hitter problem for $n$ parties, each of which possesses a single element from a universe of size $d$. However, their algorithms experience high error between the estimated frequency for the heavy hitter items and their true frequency, where the error rate is given by
$\mathcal{O}\sqrt[\leftroot{-2}\uproot{2}6]{\frac{ log (d) log(\frac{1}{\delta})}{\epsilon^2 n}}$, which does not match their    error lower bound $\Omega(\frac{1}{\sqrt{n}})$.   In contrast to \citep{hsu2012distributed},     \citep{bassily2015local}	provide the first polynomial time local $(\epsilon, 0)$-differentially private protocol for heavy hitters that has worst-case error  $\mathcal{O} ( \sqrt{\frac{ log (d) }{\epsilon^2 n}})$. They also show that using the public coin model, each user can send only one bit to the server. However, one of the main limitations of their approach is the high time complexity, where their algorithm requires a server running time of  $O(n^{5/2})$ and a user running time of $O(n^{3/2})$. In later work,  \citep{bassily2017practical}  have proposed two algorithms,  TreeHist and Bitstogram,  which require a   server running time of   $\mathcal {O}(n)$ and a user running time of $\mathcal {O}(1)$. The TreeHist algorithm is based on a noisy, compressed version of the count sketch proposed in \citep{charikar2004finding}.   
 From the practical point of view, in a concurrent work \citep{apple}, Apple has proposed the Sequence Fragment Puzzle (SFP) algorithm, a state-of-the-art sketching-based algorithm for discovering heavy hitters using local DP and an unknown dictionary. In this work, they have proven expressions for balancing the trade-offs among privacy, accuracy, transmission cost, and computation cost, allowing a trade-off of these parameters in different practical use cases. There are some other works  (\textit{e}.\textit{g}., \citep{fanti2015building}) that propose a heuristic algorithm that can be used for finding the heavy hitter with an unknown dictionary. While the work in \citep{bassily2017practical} requires public randomness and coordination between the server and users,  the authors in \citep{acharya2019hadamard} have proposed an algorithm based on Hadamard Response (HR) that is used in general for frequency estimation and does not require any public randomness, but at the cost of a per-user communication cost of $\log (d)$, while working for all privacy regime (\textit{e}.\textit{g}., $\forall \epsilon$). In contrast to  \citep{acharya2019hadamard} that trades the need for public randomness with more per-user communication cost,  \citep{acharya2019communication} proposes an algorithm that requires only $1$-bit per user while not requiring any public randomness. However, their algorithm gives an optimal error rate only at the high privacy regime, \textit{i}.\textit{e}., $\epsilon <1 $. 
 
 The previously mentioned works utilize local DP to ensure privacy, yet it is known that local DP often leads to a significant reduction in utility \citep{kairouz2014extremal,kairouz2016discrete,duchi2013local}. On the other hand, the choice of using central DP requires having a trusted server that can first collect the clean data and then perturbs it. Since in the central DP setting, noise is only applied once by a trusted server, central DP has better utility than local DP. To overcome the limitations of central DP and local DP,  \citep{zhu2020federated}  propose trie-based heavy hitters (TrieHH) algorithm that is interactive (\textit{e}.\textit{g}., multi-round algorithm)  and leverages its interactivity to achieve central DP without the need to centralize raw data while also avoiding the significant loss in
utility incurred by local differential privacy. The DP privacy guarantee of their algorithm is achieved by leveraging the randomness from the user sampling and the anonymity properties of their distributed algorithm, which make their algorithm inherently differentially private without requiring additional noise. This is different from 
the previously discussed works that are non-interactive and achieve local DP using the randomized response.   It is also different from the work in \citep{bassily2017practical} that relies on public randomness. They have also studied the trade-off between privacy and utility and shown that their algorithm can achieve good 
 utility while ensuring strong privacy guarantees, compared with the works that rely on DP, such as \citep{apple}.

\noindent \textbf{Secure Multi-party Computing}.  
Leveraging secure multiparty computing primitives is another direction for privately computing the heavy hitters without impacting the utility \citep{9519492} or requiring a large number of users as in \citep{zhu2020federated} to get reasonable utility.   The proposed protocol by \citep{9519492} for solving the
private heavy-hitter problem leverages a  lightweight cryptographic tool called incremental distributed point functions instead of using DP, which could reduce the utility. The proposed protocol relies on the assumption of having two non-colluding servers, which is one of the main limitations of this work. Additionally, it requires at least one of the two servers to not collude with any client. Apart from these limitations,   this protocol can guarantee correctness in the presence of malicious clients who can manipulate its input string to alter the protocol execution. The proposed protocol is interactive, requiring all users to participate only once in the protocol execution, where each client can send only   
a single message of size linear in the length of the input string to the servers. Similar to most works that utilize DP, the proposed protocol requires any public-key cryptographic operations except for establishing secret channels between the
parties.      

\noindent \textbf{Secure Multi-party Computing with DP.} By combining MPC and DP,  
 \citep{DP-MPC}  have proposed a heavy hitters protocol that provides high utility even for a small number of users, which
is the most challenging regime for DP  \citep{zhu2020federated}.   The proposed algorithm, in contrast to \citep{9519492}, considers the existence of only one server that wishes to compute the K-heavy hitters on the input strings of the clients.   

\subsection{Median} 
Similar to the heavy hitter problem, the works for distributed median computation are also broadly classified  from the perspective of privacy into works that leverage MPC primitives  and  DP.  

\noindent \textbf{Secure Multi-party Computing.}
As pointed out by  \citep{aggarwal2010secure}, the problem of private computing of the $k$-th ranked element on the private dataset of several parties can be solved by constructing a combinatorial circuit that is evaluated securely by the parties   (\textit{e}.\textit{g}., \citep{goldreich2019play}). 
However, the main limitation of these generic protocols is the communication overhead. In particular, for a two-party setting, where the combined data set size is $n$, and the elements of the dataset are drawn from a field of size $M$, the communication cost of this circuit-based solution is $\Omega (n \log M)$. For applications where the data size is large, these generic solutions are impractical. 
By using an interactive protocol that relies on the binary search and secure comparison using Yao’s garbled circuit,  \citep{aggarwal2010secure}  have provided the first specialized protocols for computing the $k$-th ranked element with sublinear communication and computation overhead for the two-party setting and the multi-party setting where parties in both settings are interested in knowing the $k$-th ranked element. In the two-party case, the cost of computing the $k$-th ranked element is $O(\log M \cdot\log k)$ compared to   $O(\log^2 M )$ in the multi-party setting. The number of rounds of the proposed algorithm for the two-party is logarithmic in the number of input items, whereas the number of rounds of the multi-party algorithm is logarithmic in the size of the domain of possible input values (\textit{e}.\textit{g}., $\log M$). The proposed protocol provides security against malicious parties. 
One of the main limitations of this work for the multi-party setting is that it requires lots of coordination between all pairs of parties for establishing pairwise communication channels, thus impacting its practicality.   Another practical limitation is that it is very interactive, where the number of rounds to complete the protocol scales logarithmic with the field size.    To overcome such limitations,  \citep{tueno2020secure}  have proposed efficient algorithms that leverage the client-server architecture. In this client-server setting, there are communication channels only between each client and the server, while only clients provide inputs to the computation. The rule of the server in this setting is to make their computational resources available for the computation but have no input to the computation and receive no output. By using this setting, their proposed algorithm is less interactive,  as it only requires a fixed number of rounds with the server   (\textit{e}.\textit{g}., at most four rounds) compared to $O(\log^2 M)$ for the algorithm in \citep{aggarwal2010secure}. The highest computation cost of their algorithms is  $O(\log^2 M)$.

\noindent \textbf{Differential Privacy.}
Computing the exact median value and revealing it to the clients using the algorithms proposed by \citep{goldreich2019play,aggarwal2010secure,tueno2020secure} can violate the privacy of the parties that own this median value. To overcome such a challenge,  \citep{boehler2022secure} proposes an efficient algorithm for computing a differential private median between two parties by utilizing the exponential mechanism. The proposed algorithm has a  computation complexity sublinear in the size of the data universe (\textit{e}.\textit{g}., $\log M$).   \citep{bohler2020secure+}  proposed another algorithm for private median computation in the multi-party setting while using the exponential mechanism. Their algorithm for the multi-party setting also has a computation complexity sublinear in the data size. The threat model considered in this setting is the semi-honest  (non-malicious) clients. They also  discuss  how to extend their  algorithm
to malicious clients, and implement it using the SCALE-MAMBA framework \citep{Maba}.

\noindent \textbf{Non-private.}
From the distributed optimization perspective, \citep{iutzeler2017distributed}  has proposed distributed synchronous and
asynchronous algorithms for computing median and other elements of specified ranks of the clients' data. Unlike the works in \citep{aggarwal2010secure,boehler2022secure,bohler2020secure+} that connect all nodes as a fully connected graph, this work considers a general undirected connected graph. To distributedly solve the median problem, they first design a convex optimization problem whose solution meets the median or the quantile to compute. They solve the problem using the distributed formulation of ADMM proposed by \citep{lions1979splitting,boyd2011distributed}.

\subsection{Key-Valued data}
The objective of this problem is to collect two fundamental statistics of key-value pairs, including frequency of keys and mean of values. One naive solution is to apply local DP independently at the keys and values. Since keys are categorical data, some existing DP methods  (\textit{e}.\textit{g}., \citep{erlingsson2014rappor,kairouz2014extremal})  can be applied to each key, while each   value can be  perturbed using   (\textit{e}.\textit{g}., \citep{duchi2014privacy,nguyen2016collecting}). However,  the main challenge for this naive approach of applying local DP is to achieve a good utility-privacy trade-off, since the data contains two dimensions, and a user may have multiple key-value pairs. Additionally, this naïve independent perturbation does not preserve the correlation between the keys and values. To address this challenge, \citep{8835348} proposed the first specialized LDP algorithms for this problem by modifying the Harmony randomized response-based protocol \citep{nguyen2016collecting} to better maintain the relationships
between the keys and values to improve the accuracy of statistics while still achieving local differential privacy. Their first proposed algorithm, PrivKV,  is a non-iterative (non-interactive) algorithm that is suitable for low communication cost scenarios. Additionally, they have proposed another two interactive protocols  (PrivKVM and  PrivKVM+) to iteratively improve the estimation of a key’s mean value  PrivKVM trades the communication cost with the accuracy while PrivKVM+ balances between accuracy and communication bandwidth. The main limitation of their non-interactive algorithms is the large number of rounds required to get an unbiased mean estimation and to improve the estimation of a key’s mean value. In general, their key limitations, which have also been highlighted by  \citep{gu2020pckv} include (1) A large number of rounds requires all users to be always online, thus limiting its practicality. (2) The privacy budget increases with the number of rounds. For a fixed privacy budget, the budget for each round decreases as the number of rounds increases. This decrease in per-round privacy budget increases the amount of noise added, which can negatively impact performance.
(3) Their privacy analysis lacks improved budget composition for local differential privacy  that can capture the correlation between key and value given by their algorithms. (4) Finally, their proposed random key sampling method, which is part of their algorithms, does
not work well for a large key domain. Follow-up work by  \citep{gu2020pckv}  introduced a non-interactive framework called PCKV  with a better utility-privacy trade-off that overcomes the aforementioned limitations. In particular,  they
apply an advanced sampling procedure to enhance utility over the naive random sampling done by PrivKVM. They also require only a single iteration and provide a tighter analysis of the privacy budget consumption. 
\section{Existing solutions to set queries}\label{sec_related-works-sets}
 Private set intersection/union computations have had a number of practical use cases that is large enough to garner the attention of researchers over the last two decades ~\citep{pinkas2018scalable}.  Below, we discuss a number of key approaches to solving these set query problems, mainly from the MPC community. A taxonomy of the privacy-preserving techniques used for these set queries is given in Table 2.

\begin{table}
\resizebox{\textwidth}{!}{%
\begin{tabular}{|l|l|l|}
\hline
\textbf{Query}                                  & \textbf{Privacy technique}               & \textbf{Related works}                                                                                                                                                                                                        \\ \hline
\multirow{4}{*}{Private set intersection}    & Homomorphic encryption      & \begin{tabular}[c]{@{}l@{}}\citep{huberman1999enhancing,de2010linear,meadows1986more}\\ \citep{ion2017private, freedman2016efficient} \\ \citep{chen2017fast,hazay2017scalable}\end{tabular} \\ \cline{2-3} 
                                             & Oblivious polynomial evaluation & \citep{freedman2004efficient,dachman2009efficient}                                                                                                                                                  \\ \cline{2-3} 
                                             & Oblivious transfer              & \begin{tabular}[c]{@{}l@{}}\citep{pinkas2014faster,pinkas2015phasing, rindal2017improved}\\ \citep{kolesnikov2017practical, pinkas2019spot}\end{tabular}                           \\ \cline{2-3} 
                                             & Garbled circuit                 & \citep{huang2012private, dong2013private,inbar2018efficient}                                                                                                                                        \\ \hline
\multirow{2}{*}{Private set union}           & Homomorphic encryption    & \citep{kissner2005privacy,frikken2007privacy}                                                                                                                                                       \\ \cline{2-3} 
                                             & Oblivious polynomial evaluation & \citep{kolesnikov2019scalable,jia2022shuffle}                                                                                                                                                       \\ \hline
\multirow{2}{*}{Private cardinality testing} & Homomorphic encryption  & \citep{ghosh2019communication,badrinarayanan2021multi}                                                                                                                                              \\ \cline{2-3} 
                                             & Oblivious transfer              & \citep{branco2021multiparty}                                                                                                                                                                        \\ \hline
\end{tabular}}
\vspace{0.5em}
\caption{Taxonomy of the privacy-preserving techniques used in the set queries.}
\end{table}

\subsection{Private set intersection}  \label{sec:set_intersections}
The existing approaches  for the two-party setting  include   works based on homomorphic encryption (HE) \citep{huberman1999enhancing,de2010linear,meadows1986more, ion2017private, freedman2016efficient, chen2017fast},   works  based on Oblivious Polynomial Evaluation \citep{freedman2004efficient,dachman2009efficient}, works based on  Oblivious Transfer \citep{pinkas2014faster,pinkas2015phasing, rindal2017improved, pinkas2019spot}, and works based on garbled circuit \citep{huang2012private, dong2013private}. Although these techniques are for the two-party setting, some of them were extended to the multi-party setting. Specifically,   ~\citep{kolesnikov2017practical}  have proposed oblivious programmable pseudo-random functions that are based on the idea of using oblivious transfer. Garbled  bloom filter has been used in  (\textit{e}.\textit{g}.,  \citep{inbar2018efficient}), and HE has been used in  (\textit{e}.\textit{g}., \citep{hazay2017scalable}).

\subsection{Private set union} 
 \citep{kissner2005privacy} have proposed the first protocol for the private set union, which leverages threshold additively HE and polynomial representation. Another approach \citep{frikken2007privacy} that adopts a similar technique can reduce the communication/computation complexity of  \citep{kissner2005privacy}. Instead of using polynomial representation,  \citep{davidson2017efficient} uses an inverted Bloom Filter. While the above works use public key operations, which result in increasing their computation complexities, \citep{kolesnikov2019scalable} proposed the first scalable PSU protocol using only symmetric-key techniques while using polynomial representation for computing the private set unions. However, their protocol requires repeated high-degree polynomial interpolations on the parties' datasets. To overcome such limitation,  \citep{jia2022shuffle} proposed an algorithm that relies on using data shuffling and avoids using HE and repeated operations.

\subsection{Private cardinality testing} The problem of cardinality testing has been considered in the two-party setting by \citep{ghosh2019communication, bhowmick2021apple}, and in different works for the multi-party setting by  \citep{branco2021multiparty,badrinarayanan2021multi} where these different works have developed efficient solutions in terms of the computation and communication costs while preserving the privacy of the users' data.
\begin{table}
\resizebox{\textwidth}{!}{%
\begin{tabular}{|l|l|l|}
\hline
\textbf{Query}                 & \textbf{Privacy technique}      & \textbf{Related works}                                          \\ \hline
\multirow{3}{*}{Matrix factorization} & Homomorphic encryption & \citep{liu2019privacy,chai2020secure} \\ \cline{2-3} 
                                      & MPC                    & \citep{a2}                            \\ \cline{2-3} 
                                      & DP                     & \citep{berlioz2015applying}           \\ \hline
\end{tabular}}
\vspace{0.5em}
\caption{Taxonomy of the privacy-preserving techniques for  matrix transformation.}
\end{table}

\section{Existing solutions to matrix transformation}\label{sec_related-works-matrix}  
To solve the problem in \eqref{eq-svd}, \citep{a2} proposed  an efficient  lossless  federated
SVD solution over billion-scale data called FedSVD ensures the accuracy of the SVD computation is not impacted. This is guaranteed by avoiding using DP methods; instead, they rely on masking their data in a way such that the masks are canceled out when the response from the different parties is aggregated by the server. Thus,  this approach guarantees the same performance as the centralized case where all the data are located in one place. \citep{liu2019privacy} have proposed an algorithm that uses additive HE. On the other hand,  \citep{chai2020secure, berlioz2015applying} have proposed distributed privacy-preserving algorithms for recommendation systems that rely on matrix factorization. The proposed algorithm by  \citep{chai2020secure} is based on HE, while the one proposed by \citep{berlioz2015applying} leverages differential privacy. The  taxonomy of the privacy-preserving techniques used for the set queries is summarized  in Table 3. 

{\section{Challenges and Open Opportunities}

\subsection{Algorithmic security and privacy}
In the previous Sections 4-6, we presented a number of privacy-preserving approaches to compute the FA queries. However, unlike FL, there does not exist a single common framework or algorithm for privately computing a diverse number of queries. A unifying approach to evaluate FA queries without leaking unnecessary information is an open question of great importance for deploying FA systems, as it will allow them the flexibility to deal with a wide range of queries.
Note that if the target is to solve the query while disregarding privacy, then a number of queries discussed earlier can be computed and then used to derive answers for other queries. For example, the mode, mean and median statistical queries can all be computed by first computing the FA histogram query and then deriving the target answers (\textit{e}.\textit{g}. median) from it. This, however, leaks unnecessary information to the querer beyond the intended goal.

One solution to address this information leakage is to employ secure enclaves~\citep{costan2016intel} at the querer to isolate a code execution and memory in a trusted environment where the code can be attested and verified while keeping its state a secret until it publishes an output. Using this in our previous example, the querer can run a code to aggregate the histogram and then extract the required target query from it. Although secure enclaves can theoretically address the security challenges arising from using a non-specialized analytics algorithm, current secure enclave models are only limited to CPU resources and provide limited memory resources, which limits their potential universal deployment. 

With these limitations, it remains an open problem when and how much to make use of these trusted secure enclaves in the logic for computing the target query, and whether there exists a universal approach to securely and privately computes federated analytics queries that does not need to use secure enclaves.

\subsection{Robustness to system failures}\label{sec:robustness_challenge}
The quality of computed analytics in a federated analytics system can be prone to performance degradation due to a number of malicious or non-malicious system failures. Malicious failures can arise due to attempts by some system parties to alter their data or responses in order to either degrade the system performance or targets its deviation towards a premeditated result. In addition to malicious failures, the distributed nature of federated analytics and its reliance on parties that are not co-owned can cause it to suffer from party dropout or straggling which can potentially happen during the execution of the federated analytics algorithm. The use of privacy-preserving mechanisms in federated analytics such as secure aggregation~\citep{Bonawitz2016PracticalSA} as well as other MPC protocols, can hinder the detection or recovery from these malicious or non-malicious faults. How to make federated analytics robust to such failures without giving up any or little privacy is an interesting open problem in the area.

Although a universal solution for robustness in federated analytics is still open, there exist some approaches for handling failures in federated learning that can lend themselves easily to the federated analytics framework. The non-malicious failure of clients was an overarching limitation of the vanilla secure aggregation protocol~\citep{Bonawitz2016PracticalSA}. While the protocol design was inherently able to recover from these failures and compute the sum (mean) from the surviving clients, a huge recovery cost is incurred that is can grow quadratically with the number of clients. Recent advances~\cite{secagg_kadhe2020fastsecagg,so2021lightsecagg} have proposed more efficient approaches for designing secure aggregation keys that allow for a more efficient recovery. These techniques lend themselves to algorithms that rely on aggregating from all clients simultaneously. Some federated queries, however, require structured responses where a particular subset of clients need to be active in each round. In this case, recovering the aggregate response from the surviving clients may be useless in some cases, and more sophisticated secure aggregation protocols are in great need. For example, one simple method would be checking if the subset of surviving clients does not satisfy particular properties, and if so, abandon the aggregation over this subset of clients in this round.

For malicious failures that try to poison a client's dataset, data sanitization~\citep{cretu2008casting,steinhardt2017certified} and anomaly-detection~\citep{blanchard2017machine} techniques, which aim to detect or remove anomalous data, have typically been used to address this. However, these techniques typically rely on access to some subset of the clients' data at the server or the availability of data that is sampled from the same distribution, which makes them incompatible with privacy-preserving approaches employed in federated analytics. It remains an open problem whether we can use these failure mitigation techniques in federated analytics without giving up privacy or if new defense approaches need to be developed to address malicious failures in federated analytics.

\subsection{Participation incentive mechanisms}
In parallel to the development of efficient and secure approaches for federated analytics, developing appropriate mechanisms to incentivize participation is a critical open question for federated analytics systems. This is particularly important in scenarios where the data owners are competitive entities such as financial institutions or enterprises, where the default strategy is not to collaborate with other competitors. Forms of incentive in the cross-silo setting can be regulatory by a governing entity (for example, the FDIC wants to detect fraudulent activity across different banks~\citep{elkordysecure}), or for shared operational stability, by jointly computing the salary quantiles across a cohort of companies~\citep{kenthapadi2017bringing}.
In the case of cross-device (individual) clients, incentives can include provided services, and/or monetary gain. From a service perspective, federated analytics promises users potential improvement in the quality of their service experience, \textit{e}.\textit{g}., a higher accuracy word predictor in Gboard or better estimation of travel times in navigation applications. In other scenarios, the incentive can be individual welfare, similar to the contact tracing analytics performed using private set intersections during the COVID-19 pandemic. 

In either cross-silo or cross-device, a central challenge is balancing incentive with the heterogeneity of data and contribution (\textit{e}.\textit{g}., in terms of the data size). To address this, careful design should be taken into account to ensure clients with more data are not discouraged due to the non-proportionality of the incentives to their contributions, as well as, not pushing away clients with less data by not implementing worthwhile incentives.

\subsection{Decentralized and trust}
Our discussions so far always considered a central querier that poses intermediate questions to the clients and aggregates their responses in order to arrive at the query answer (this can be in one-shot or iteratively). Such a model makes sense for queries where the question implies an authoritative entity (for fraud detection for instance) or a large company (for product analytics) is asking the query. However, for a population of clients that wish to collaboratively learn a property of their joint dataset, handling the query computation distributively can be more desirable. The key idea of decentralized analytics is to rely on peer-to-peer communications between the clients to answer the query, while still maintaining the privacy and security of exchanged information about the local datasets. Computing decentralized analytics can find application in scenarios such as the evaluation of trained models that are stored on the blockchain~\citep{shayan2020biscotti} or to crowd-source the computation of percentiles (\textit{e}.\textit{g}., median) of employee salaries of the technology sectors without the pre-requisite of having the parent companies agree to perform this federated computation.

There has been a wide array of works in MPC that develop decentralized solutions for secure computation, particularly for private set intersection problems (see \S\ref{sec:set_intersections}).
However, such solutions assume that the communication graph of clients is fully-connected and undirected. This can lead to inefficient protocols, particularly as the number of parties increases. Furthermore, sparse and directed communication graphs can model more diverse scenarios, for instance, when the clients are not co-located or when communication goes in a single direction (\textit{e}.\textit{g}., due to different social network connection tiers). 

An interesting aspect of decentralized federated analytics is its decreased robustness to system failures (see the discussion in \S\ref{sec:robustness_challenge}) due to the absence of a centralized entity that can potentially filter out malicious contributions or recover the system in the case of party drops. The design of incentive mechanisms for participation in a decentralized scenario is also a critical open research direction, as coordinating incentives is also impacted by the absence of a central coordinator.

One recent promising approach to address decentralized analytics challenges is to use blockchains to keep track of intermediate updates and verify that intermediate clients in the communication graph do not act maliciously during the aggregation of updates. The Biscotti framework~\citep{shayan2020biscotti} in the context of federated learning can be easily extended to mechanisms that rely on iterative updates and secure aggregation. In Biscotti, the blockchain ledger uses verifiable random functions to ensure that the aggregation contributed by a user is truly the resultant of the stored encoded intermediate updates. It also uses DP to ensure the privacy of these stored encodings. An adaptation of a blockchain solution for decentralized federated analytics can lead to more flexible algorithms that are crowd-operated without the requirement to trust a centralized aggregator/querier entity.
}
\subsection{Cross-silo federated analytics on the cloud}
In previous sections, we assume that FA clients own their data and process the data in local and trusted environments when responding to a query. 
However, in real-world deployments, instead of maintaining local data centers and keeping the data on the local side, FA clients typically would use third-party public cloud services such as Microsoft Azure, Google Cloud, Amazon Web Services, IBM Cloud, and Alibaba Cloud, to store and process their data. Outsourcing data to such third-party clouds has emerged as the de facto model for data storage and processing for numerous benefits, such as improved availability, lower cost, and improved service.



{Using clouds in an FA system, however, poses additional security and privacy challenges due to the untrusted nature of public clouds. A public cloud may be curious and wish to learn some information about the data of the FA clients. To protect data from such adversarial clouds, FA clients can use two classes of solutions for secure data outsourcing.
The first is called \textit{single cloud-based solutions}, in which clients encrypts their data and use a single cloud to store the data. The second is called \textit{multi-cloud-based solutions}, in which a client partitions their data into several parts, \textit{e}.\textit{g}., secret-sharing shares, and stores those parts in different clouds so that no single cloud can get the complete data.}

In the following subsections, we will discuss solutions in the literature that address security and privacy challenges in the two aforementioned outsourcing settings. {We will use the set intersection query as the running example throughout our discussions, since it is difficult, complex, and important in query processing, and most existing works focus on this type of query.}

\subsubsection{Single cloud-based  solutions.} 
In single cloud-based solutions, clients encrypt their local data and use a single cloud to store the data. 
\citep{VDPSI,abadi2017efficient,kamara2014scaling,kerschbaum2012collusion, liu2014encrypted,qiu2015identity, abadimulti, zhang2017server}
 allow clients to outsource their private datasets and process Private set Intersection (PSI) tasks without downloading the datasets. \citep{abadi2017efficient} ensures that the cloud can only compute set intersection after obtaining the permission of all the clients, and the computation results will be protected from the cloud. 
\citep{VDPSI,kamara2014scaling} provided a watermark-based verification approach for queries over outsourced encrypted datasets. \citep{VDPSI} can also detect malicious cloud (\textit{i}.\textit{e}., an adversarial cloud that may tamper the data stored on it) by inserting secret values in the real datasets to the cloud each time to process a PSI query. 
By checking whether the result set contains the secret values, the clients will know whether the query result is correct or not. 
\citep{kerschbaum2012collusion} shares secrets between the cloud and the clients to pre-process datasets when outsourcing the datasets. This approach is collusion-resistant if one client and the public cloud collude. However, it requires a client to encrypt the datasets with different encryption keys for set intersections with different clients. 
\citep{liu2014encrypted} delegates PSI computation over randomized datasets to a cloud. Each client computes the hash value of its dataset using a general-purpose hash function, then randomizes each hashed data with a random integer.
\citep{qiu2015identity} applied fine-grained authorization that enables the cloud to perform queries without leaking any data. 
When a client A asks for a matching request with another client B, A first negotiates a token with B so that A can delegate the computation over the outsourced encrypted datasets to the cloud server, and such operations require a trusted third party to generate a token on behalf of the clients. 

With the exception of~\citep{kamara2014scaling}, the aforementioned techniques have quadratic/exponential complexity or use expensive cryptographic techniques~\citep{qiu2015identity}, and as a result, do not support large-sized datasets at the FA clients. While \citep{kamara2014scaling} scales better, it does not support aggregation, and, moreover, reveals which item is in the intersection set. Fed-K-PSI \citep{elkordysecure} is a different variant of the server-based federated PSI. Each record on the client's side is represented by a key-value pair,  and the server is the entity that is interested in knowing the set of identifiers that appears associated with the same value at least $K$ times. One of the main components of Fed-K-PSI  is the   secure aggregation  protocol that has  been widely used in FL setting  \citep{so2021lightsecagg,jahani2022swiftagg,elkordy2022much,9712310,Bonawitz2016PracticalSA}

\subsubsection{Multi-cloud-based solutions.}
In multi-cloud-based solutions, a client partitions his/her local data into several parts, \textit{i}.,\textit{e}., shares, and stores each share at different clouds. Each cloud only has partial information, thus a single cloud can not learn actual dataset~\citep{bater2017smcql,volgushev2019conclave,prism,corrigan2017prio}. 
To partition data into shares, Shamir's secret-sharing~\citep{shamir79} is the most widely-used technique.

Prio~\citep{corrigan2017prio} is a privacy-preserving system for collecting statistics that allows multiple clients to upload their data in shares to multiple clouds, and these clouds execute only aggregation operations -- count, max/min/median. Prio allows servers to verify the data they receive before storing it at their end. 
However, Prio only offers a mechanism for confirming the maximum number if the maximum number is known while does not provide any mechanism to compute the maximum/minimum number.  {Concalve}~\citep{volgushev2019conclave} is an additive sharing-based system that allows to execute SQL queries over multiple clients. Conclave allows partitioning the computation such that parts of the computation can be executed at the client over cleartext and the remaining parts can be executed over additive shares. For example, a join query with selection can be partitioned such that the selection condition can be executed at clients, and then the clients create additive shares of the data that qualifies the selection condition. On the additive shares, a join query over the additive shared data belonging to multiple clients can be executed. 
Two other systems similar to Conclave are  {Senate}~\citep{poddar2021senate}, which allows collaborative SQL processing among multiple clients without using the cloud, and {SMCQL}~\citep{bater2017smcql}, which is a garbled circuit based system supporting PSI via join and aggregation operations. However, these systems are inefficient when processing large datasets due to either potential memory outage and/or multiple communication rounds in the cloud. For example, SMCQL takes $\approx$ 23 hours over 23M values, while Conclave takes 8 mins over 4M values. Furthermore, to execute PSI via join operation, Conclave needs to reveal the joining column in cleartext to a trusted third party.
Helen~\citep{zheng2019helen} and
Cerebro~\citep{zheng2021cerebro} are two recent systems that perform collaborative machine learning tasks without using the cloud. 
Another recent system for executing queries in the multi-cloud-based is Prism~\citep{prism}. Prism uses both additive shares to support Private Set Intersection (PSI)/Union (PSU) operations and multiplicative shares to offer aggregation. 
Furthermore, Prism~\citep{prism} is able to support query executions over large datasets and multiple clients. To securely execute a computation, Prism needs at most three non-colluding cloud servers. Prism does not require communication among servers during/after/before the computation, and, consequently, is able to support PSI/PSU over 20 million values in 8 seconds. Furthermore, Prism is the only system that supports result verification operations. 

\section{Conclusion}
In this article, we provide an overview of federated analytics, a privacy-preserving paradigm to solve queries over distributed data owned by multiple clients. We discussed the unique properties of federated analytics and how it relates to FL. We also provide a proposed taxonomy for different classes of queries in federated analytics and a survey of existing solutions in classical areas of distributed computing and secure computation. 
{Finally, we discussed several challenges and open directions for the application and deployment of FA systems at scale. Addressing these challenges can help bring FA systems closer to being deployed in more practical scenarios to answer a wider range of queries.}

\section{Acknowledgements}
This work is supported by NSF grants CCF-1763673, CNS-2002874, Defense Advanced Research Projects Agency (DARPA) under Contract No. FASTNICS HR001120C0088 and HR001120C0160, 
ARO grant W911NF-22-1-0165, and gifts from Intel, Qualcomm, and Cisco.

\printbibliography

\end{document}